%% file: GraVoS.tex
\documentclass[10pt,twocolumn,letterpaper]{article}

\usepackage[pagenumbers]{cvpr} 

\usepackage{graphicx}
\usepackage{amsmath}
\usepackage{amssymb}
\usepackage{booktabs}

\makeatletter
\@namedef{ver@everyshi.sty}{}
\makeatother
\usepackage{tikz}

\usepackage{tikz}
\usepackage{color}
\usepackage{comment}

\usepackage[utf8]{inputenc} 
\usepackage[T1]{fontenc}    
\usepackage{url}            
\usepackage{booktabs}       
\usepackage{amsfonts}       
\usepackage{nicefrac}       
\usepackage{microtype}      
\usepackage{xcolor}         

\newcommand*{\ShowNotes}{}
\definecolor{darkred}{rgb}{0.7,0.1,0.1}
\definecolor{darkgreen}{rgb}{0.1,0.7,0.1}
\definecolor{dblue}{rgb}{0.2,0.2,0.8}
\definecolor{maroon}{rgb}{0.76,.13,.28}
\definecolor{burntorange}{rgb}{0.81,.33,0}
\definecolor{cyan}{rgb}{0.0,0.7,0.94}
\definecolor{salmon}{rgb}{0.99,0.51,0.46}
\definecolor{green}{rgb}{0.03,0.91,0.43}

\ifdefined\ShowNotes
  \newcommand{\colornote}[3]{{\color{#1}\bf{#2: #3}\normalfont}}
\else
  \newcommand{\colornote}[3]{}
\fi

\newcommand*\cyan{\color{cyan}}
\newcommand*\salmon{\color{salmon}}
\newcommand*\green{\color{green}}

\usepackage{sg-ibs-macros} 
\usepackage{subcaption}

\usepackage[pagebackref,breaklinks,colorlinks]{hyperref}

\usepackage[capitalize]{cleveref}
\crefname{section}{Sec.}{Secs.}
\Crefname{section}{Section}{Sections}
\Crefname{table}{Table}{Tables}
\crefname{table}{Tab.}{Tabs.}

\def\cvprPaperID{499} 
\def\confName{CVPR}
\def\confYear{2023}

\begin{document}

\teaser{
    \centering
    \includegraphics[width=0.89\linewidth]{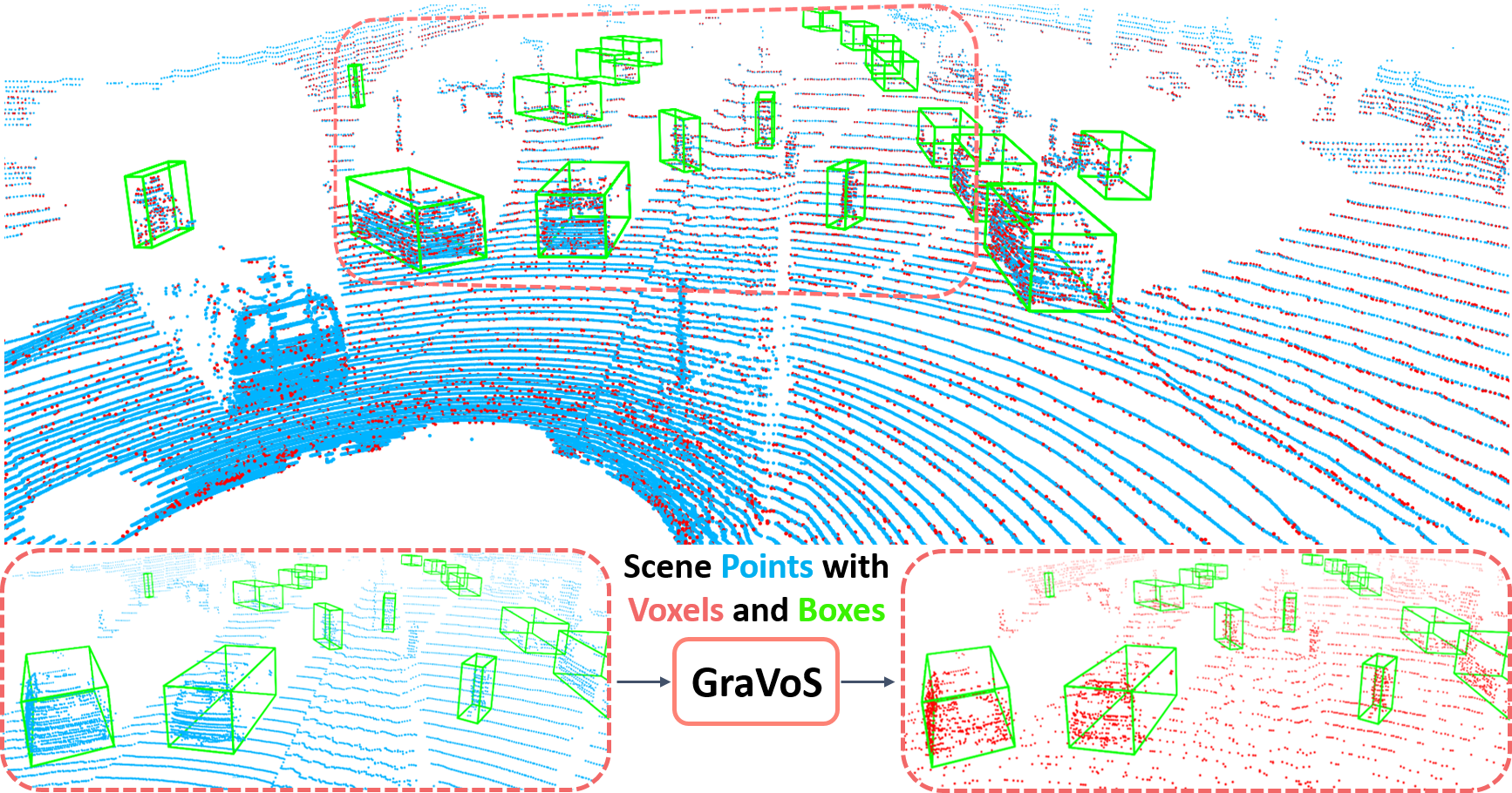}
    \caption{\textbf{GraVoS for 3D object detection}. 
    Given a 3D {\cyan point cloud (cyan)} and its associated voxels, we propose a   method that selects a subset of network-dependent {\salmon meaningful voxels (salmon)}.
    Our method selects most of the voxels of the challenging classes, those with relatively few training instances (such as the {\em Cyclists} and {\em Pedestrians}), less from the prevalent classes (e.g. {\em Cars}) and very few from the background.
    It is shown that considering only this subset improves the performance of numerous SoTA voxel-based detectors.
    }
    \label{fig:Teaser}
}

\title{GraVoS: Voxel Selection for 3D Point-Cloud Detection}

\author{Oren Shrout\\
Technion, Israel\\
{\tt\small shrout.oren@gmail.com}
\and
Yizhak Ben-Shabat\\
Technion, Israel \& ANU, Australia\\
{\tt\small sitzikbs@technion.ac.il}
\and
Ayellet Tal\\
Technion, Israel\\
{\tt\small ayellet@ee.technion.ac.il}\\
}

\maketitle
\thispagestyle{empty}

\begin{abstract}

3D object detection within large 3D scenes is challenging not only due to the sparsity and irregularity of 3D point clouds, but also due to both the extreme foreground-background scene imbalance  and  class imbalance.
A common approach is to add ground-truth objects from other scenes.
Differently, we propose to modify the scenes by removing elements (voxels), rather than adding ones.
Our approach selects the "meaningful" voxels, in a manner that addresses both types of dataset imbalance.
The approach is general and can be applied to any voxel-based detector, yet the meaningfulness of a voxel is network-dependent.
Our voxel selection is shown to improve the performance of several prominent 3D detection methods. 
\end{abstract}

\section{Introduction}

3D object detection has gained an increasing importance in both industry and academia due to its wide range of applications, including autonomous driving and robotics \cite{kitti2,thrun2006stanley}. 
LiDAR sensors are the de-facto standard acquisition devices for capturing 3D outdoor scenes (\figref{fig:Teaser}). 
They produce sparse and irregular 3D point clouds, which are used in a variety of scene perception and understanding tasks.

There are three prominent challenges in outdoor point cloud datasets for detection.
The first is the small size of the datasets, in terms of the number of scenes.
The second is  the large number of points in the scene vs. the small number of points on the training examples (objects).
A single scene might contain hundreds of thousands, or even millions, of points but only a handful of objects.
The third is class imbalance, where some classes might contain significantly more instances than the other classes.
This often results in lower predictive accuracy for the infrequent classes~\cite{johnson2019survey,oksuz2020imbalance}.

To handle the first challenge, it was proposed
 to enrich the dataset during training with global operations, applied to the whole point cloud, such as rotation along the Z-axis, random flips along the X-axis, and coordinate scaling~\cite{second,yang2018pixor,zhou2018voxelnet}.
 We note that most methods that solve the second and the third challenges indirectly also solve this first challenge.

To solve the second challenge, it is suggested to augment the scene  by local operations, applied to points belonging to individual objects~\cite{cheng2020improving,choi2021part}.
Local operations include random point drop out, frustum drop out, additive noise added to the object, intra-object mixing, and swapping regions between different objects.

To solve the third challenge, as well as the second one, \cite{pvrcnn,pointrcnn,parta2,second} propose to add ground-truth objects from different point-clouds to the scene, for training.
This type of augmentation indeed mitigates the imbalance.
However, it does not take into account the network architecture, though
Reuse \etal~\cite{reuse2021ambiguity} show, through a series of experiments, that both local and global data augmentation for 3D object detection
strongly depends not only on the dataset, but also on the network architecture.
Thus, network-dependent augmentations are beneficial.

We present a novel network-dependent data modification approach that addresses the latter two challenges.
(We address the first challenge indirectly, similarly to~\cite{pvrcnn,pointrcnn,parta2}.)
The key idea is to learn a subset of elements of the scene, which are meaningful for object detection.
Inline with~\cite{reuse2021ambiguity}'s observation, meaningfulness is defined in the context of the network and not only of the scene.
Considering only this subset as input will allow us not only to decrease the number of elements in the scene, but also to increase the class balance within a given scene.

To realize this idea, we focus on SoTA detection networks that transform point clouds into voxels as a first step in their pipeline.
The main reason behind transferring the input to voxels is reducing the size of the input, as only the occupied voxels are then processed.
This enables these systems to work on extremely large scenes.
Obviously, another benefit of voxels is the ability to impose structure on the input.

Generally speaking, in this voxel-based setup, meaningful voxels are those for which the model "struggles" to locate the objects.
Hence, the gradients play a major role in determining the meaningful voxels.
Our approach is thus termed {\em Gradient-based Voxel Selection (GraVoS)}.

We show that when focusing only on the meaningful voxels and removing the non-informative ones, most of the discarded voxels belong to the scene background.
Few of the removed voxels are associated with the prevalent classes and almost none are associated with the non-prevalent classes. 
Our strategy may be contrasted with {\em dropout} augmentation techniques, which reduce the number of elements randomly~\cite{cheng2020improving,choi2021part}.

Our distribution balancing is demonstrated in \figref{fig:Teaser}.
Given a {\cyan point cloud (cyan)}, our selected subset of the {\salmon meaningful voxels (salmon)} contains significantly few points from the background (the points are voxel centers here), more points on the objects that belong to the {\em Car} class (the most prevalent class), and almost all the points on the objects that belong the {\em Pedestrian} and the {\em Cyclist} classes.

Our method is general and may be applied to different voxel-based networks.
Furthermore, it comes at no additional inference time cost.
We show results on four SoTA networks: 
SECOND~\cite{second}, 
Part-$A^2$~\cite{parta2}, 
Voxel R-CNN~\cite{voxelrcnn}, 
and CenterPoint~\cite{yin2021center} on the well-established KITTI dataset~\cite{kitti1,kitti2}.
The performance of all networks improves, in particular when considering the difficult categories of {\em Pedestrian} and {\em Cyclist}.
For instance, the performance of~\cite{parta2} on the benchmark's moderate subset improves by $2.32\%$ \& $1.15\%$  for the non-prevalent classes ({\em Cyclist} and {\em Pedestrian}), which constitutes an error reduction of $8.20\%$ \& $2.77\%$, respectively.

Hence, the main contributions of this paper are: 
\begin{itemize}
    \item 
    A novel \& generic "meaningful" voxel selection method, called {\em Gradient-based Voxel Selection}.
    \item 
    A training procedure that uses the selected voxels to improve 3D detection without additional data. 
    This procedure combines information from different stages of the model's training.
    \item
    Demonstrating improved performance of four voxel-based SoTA detection methods, successfully coping  both with the inherent class imbalance and with the foreground-background imbalance.
\end{itemize}

\begin{figure*}[t]
    \centering
    \includegraphics[width=0.9\linewidth]{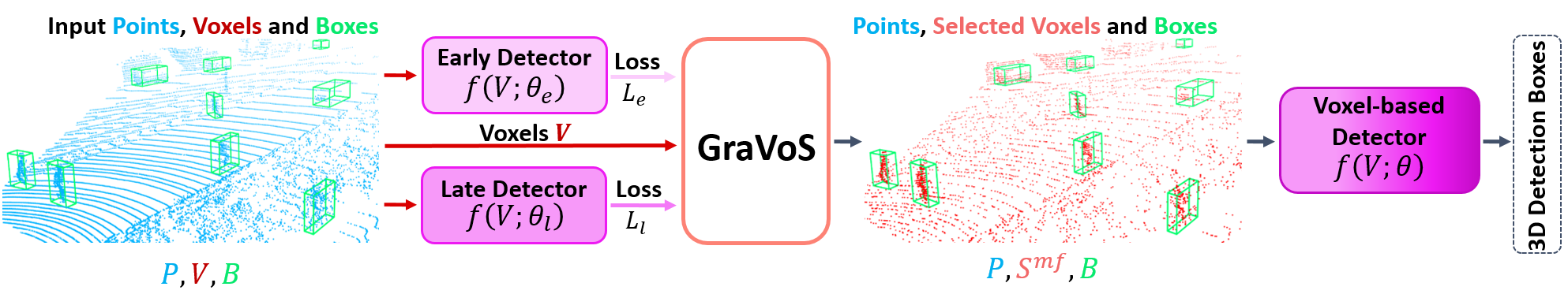}
    \caption{\textbf{Training with GraVoS}. 
    An input {\cyan point cloud (cyan)} is voxelized and fed into a pre-trained voxel-based detector at two different training stages, early and late (with frozen weights).
    These detectors' losses are computed and are the input of GraVoS, which performs voxel selection. 
    The {\salmon selected voxels (salmon)} are then fed into the late detector, initializing its weights where it left off ($\theta_l$) and continuing training using the selected voxels exclusively.
    Here, $f(\cdot)$ is a voxel-based network.
    }
    \label{fig:incorporating_gravos}
\end{figure*}

\section{Related Work}

\noindent{\bf 3D Object Detection.}
Object detection methods aim at localizing objects in a given scene and classifying them.
3D detection methods can be categorized into  grid-based~\cite{chen2019fast,voxelrcnn,pointpillars,parta2,second,yang2018hdnet,yang2018pixor,ye2020hvnet,yin2021center,zheng2021cia,zheng2021se,zhou2018voxelnet} and point-based methods~\cite{pvrcnn,pointrcnn,shi2020point,yang20203dssd,yang2019std}.
See~\cite{guo2020deep} for an excellent survey on deep learning for point clouds in general and for detection in particular. 

Grid-based methods first transform the given point cloud into a regular representation, either voxels~\cite{chen2019fast,voxelrcnn,pointpillars,parta2,second,ye2020hvnet,yin2021center,zheng2021cia,zheng2021se,zhou2018voxelnet} or a 2D Bird-Eye View (BEV)~\cite{yang2018hdnet,yang2018pixor}.
This enables processing using 3D or 2D 
CNNs, respectively.
The resulting voxels, however, have a very sparse spatial distribution~\cite{second}.
To handle sparsity, sparse convolutions~\cite{sparseconv,graham2017submanifold} have been proposed.
Modifications of the sparse convolution were proposed by~\cite{second} for efficient feature extraction and by~\cite{voxelrcnn,pvrcnn,parta2} for efficiently generating box proposals.
These grid-based approaches provide efficient and accurate solutions, but are limited when the data is imbalanced.

For point-based approaches, point set abstractions are learned directly from the raw point cloud, by utilizing PointNet-like architectures~\cite{qi2017pointnet,qi2017pointnet++}.
For instance, PointRCNN~\cite{pointrcnn} uses the PointNet backbone to generate proposals directly from the point cloud for a refinement stage.
STD~\cite{yang2019std} further improves the refinement stage by densifying the sparse proposal. 
A fusion sampling strategy of~\cite{yang20203dssd} takes into account the distances between the points in both coordinate and feature spaces. 
While point-based methods are quantization-free and have flexible receptive fields, they suffer from GPU memory limitations and from a high computation time  when processing large point clouds.

\noindent
{\bf Augmentation methods for 3D detection.}
Since the size of the available training data is limited, most SoTA 3D detection methods use a data augmentation protocol to alleviate the overfitting problem~\cite{chen2019fast,voxelrcnn,pvrcnn,pointrcnn,parta2,second,yang20203dssd,ye2020hvnet,zhou2018voxelnet}. 
Generally speaking, global operations are applied to the whole point cloud scene, including 
random coordinate scaling,
random flips along the $X$-axis, 
and random rotations around the $Z$-axis.
Despite their benefits, these augmentations do not address the challenges of data imbalance.
In~\cite{second} it is proposed to randomly insert cropped ground-truth objects  from different scenes of the training data, while avoiding overlapping objects. 

Recent works suggest to augment the training scenes also by local operations on individual objects.
These operations can be subdivided into two categories: subtractive and additive. 
For the subtractive operations, it is suggested in~\cite{cheng2020improving}  to apply random points or frustum drop out. 
For the additive operations, it is proposed in~\cite{cheng2020improving} to add noise to objects' points and in ~\cite{choi2021part} to also apply mixing or swapping parts between different objects in the same class.

\noindent
Alternative methods including bootstrapping, or {\em hard negative mining}, have been used to handle data 
imbalance~\cite{mitra2020multiview ,sung1996learning,shrivastava2016training,ren2018learning}.
The key idea is to gradually add examples to the training set, for which the detector classifier gets false positives.
This is usually done iteratively, where
the detector is first trained on the whole training set
and then the training set is updated based on the detector's new false positives.
Differently from these approaches, our training set is not explicitly updated using the ground truth information.

\section{Gradient-based Voxel Selection (GraVoS)}
\label{sec:GraVoS}

\begin{figure*}[t]
    \centering
    \includegraphics[width=0.9\linewidth]{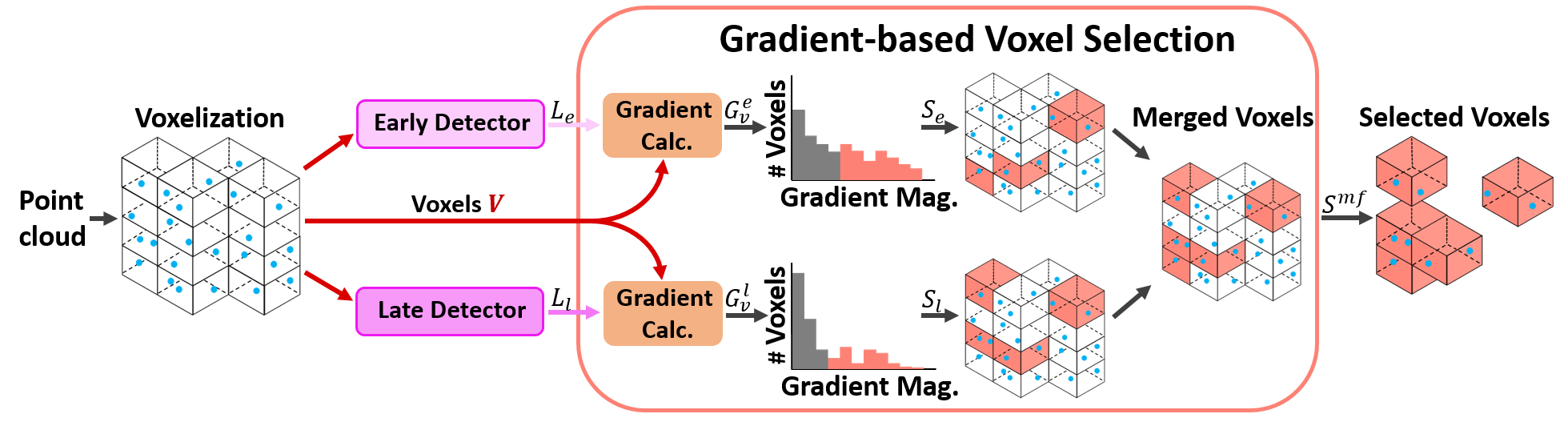}
    \caption{\textbf{GraVoS Module.} 
    The voxelized point cloud is fed into the GraVoS module and the pre-trained detector (at two training stages). 
    The detectors' losses are computed and fed into the GraVoS module. 
    These losses are used to compute the gradient magnitude at each voxel.
    For each detector stage the voxels are selected based on their gradients' magnitude.
    The selected voxels  (highlighted in {\salmon salmon}) from the two stages are then merged to form the final selected subset of voxels, $S^{mf}$.
    }
    \label{fig:gravos_illustration}
\end{figure*}

We proposes a novel subtractive approach, which provides a further boost in performance to voxel-based 3D detection methods.
We focus on two properties that make typical scenes challenging for detection systems: (1)~foreground-background imbalance and (2) class imbalance.

Given an input point-cloud of an outdoor scene, 3D detectors aim to localize objects and classify them.
We concentrate on voxel-based detectors, since they are beneficial in handling  scenes with a very large number of points.
Typically, these detectors start by converting the cloud into a voxel grid.
Then, non-occupied voxels are removed and the points in every occupied voxel are grouped.
To address the varying number of points in the voxels, random sampling of points is applied to each voxel.
These voxels are then fed into a detection network, which is usually a sparse 3D CNN, followed by
a region proposal network (RPN).
The detector outputs both bounding box proposals and class predictions.
While this approach exhibits good performance for the prevalent classes,
it might deteriorate for other classes.

\vspace{0.02in}
\noindent
\textbf{Training  with GraVoS.}
We propose to add a selection component to the above general approach.
The goal is to select the "meaningful" voxels and to remove the less meaningful ones from the scene, in a manner that addresses the two imbalance challenges discussed above.
Meaningfulness is indicated by the magnitude of the gradients, as
the gradients encapsulate information regarding the detector's success.
Selecting the voxels with high magnitudes puts more emphasis on the voxels that determine the locations of the objects.

\figref{fig:incorporating_gravos} illustrates the training of an existing 3D voxel-based detector with our {\em Gradient-based Voxel Selection (GraVoS)} module.
First, a voxel detector $f(\cdot)$ is trained on the dataset  without any modification.
Its parameters are stored at two different training stages -- early stage and late stage, $\theta_{e}$ and $\theta_{l}$, respectively.
Then, the voxels are fed into these pre-trained detectors,  $f(V; \theta_e)$,  $f(V; \theta_{l})$, separately.
Each detector's location loss is fed into our GraVoS module, along with the computational graph and the voxelized scene.
Within the GraVoS module, the meaningful voxels of the voxelized scene are found. 
These become the input to a new copy of the late stage detector $f(V; \theta)$, which is further trained using {\em only} this refined subset.

We use two different training stages since they provide complementary information regarding voxel meaningfulness. 
The late stage detector assigns higher values to meaningful and unique voxels and lower values to meaningful voxels in repetitive and easy to learn features. 
Conversely, in the early stage detector, these "easy" features still maintain a high gradient magnitude. 
Though learned early on, they are essential for recognizing the objects.
Hence, merging the two sets is beneficial, as shown in the ablation study.

\vspace{0.02in}
\noindent\textbf{GraVoS Module.}
GraVoS aims at selecting the meaningful voxels and discarding the less informative ones.
Its structure is illustrated in \figref{fig:gravos_illustration}.
A voxelized grid of the point cloud scene ({\cyan cyan}) and the detector location losses from the early and the late stages, as well as the computational graphs, are fed into GraVoS.
Then, for each detector's loss, the gradients' magnitude are calculated and the meaningful voxels are found ({\salmon salmon}).
Finally, the voxels that pass the threshold for each detector's stage are merged, to create the selected voxel set.
We elaborate on the selection process hereafter.

The magnitudes of the gradients of the losses w.r.t. each input point (each voxel contains its points) is computed.
These are then aggregated  for all the  points in the voxel, to get a scalar value per voxel.
This value  will be later used as an informativeness measure.
Formally, let $L_{e}, L_{l}$ denote the computed location losses for the early and the late stages, respectively.
(These losses are associated with the specific detector and are defined accordingly; see the supplementary for the exact definitions.)
The gradient magnitude per voxel, $v_{i}$, is computed as:
\begin{equation}
G_{v_i}^e = \frac{1}{n_{i}}\sum_{p^i_j \in v_i}\left\lVert\frac{\partial L_e}{\partial p^i_j}\right\rVert, ~
G_{v_i}^l = \frac{1}{n_{i}}\sum_{p^i_j \in v_i}\left\lVert\frac{\partial L_l}{\partial p^i_j}\right\rVert,
\end{equation}
where $n_i$ is the number of points, $p^i_j$, in voxel $v_{i}$.

For each detector stage, early and late, we use the gradient magnitude to create a subset of  meaningful voxels $S_e$ and $S_l$, respectively.
For the late stage detector we assign the voxels with the {\em top-k} magnitude values $G^l_{v_i}$ to  $S_l$.
For the early stage detector we assign voxels with magnitude values $G^e_{v_i}$ larger than the mean $\overline{G^e_{v_i}}$ to  $S_e$.
The different threshold mechanisms are due to the fact that at the early stages the high gradients are noisy, thus we should not consider only the largest gradients.
We show the benefit of the different mechanisms in the ablation study.
Formally, the subsets are defined in \Crefrange{eq:voxel_subsets_start}{eq:voxel_subsets_end}: 
\begin{alignat}{2}
    \label{eq:voxel_subsets_start}
    &S_e = \{v_i ~ \lvert ~ G^e_{v_i} \geq \overline{G^e_{v_i}} \}, \\
   &S_l = \{v_i ~ \lvert ~ G^l_{v_i} \in \text{\em{top-k}}(G^l_{v_i}) \}.
    \label{eq:voxel_subsets_end}
\end{alignat}

The parameter $k$ gives control over the percentage of voxels that are considered meaningful.
It is selected based on two variables, (1)~$n_{vs}$, the number of meaningful voxels in the final selected set ($S^{mf})$ and (2)~$\nu_{idr}$, the intra-detector ratio between the late and the early detectors.
It is calculated as $k = \nu_{idr} \cdot n_{vs}$.
Both $n_{vs}$ and $\nu_{idr}$ are hyper parameters.
(In practice, they are set to $80\%$ of the number of input voxels and to $50/80$, respectively).

Next, we merge the subsets above by a union operator, to form the final meaningful voxels  subset:
\begin{equation}
    \label{eq:voxel_subsets_start1}
    S^{mf} = S_l \cup S_e.
\end{equation}

\input{figures/gradients_fig}

Finally, the selected voxels of $S^{mf}$ are fed into the pre-trained detector $f(V;\theta)$, which is fine tuned for several epochs using the detector's original losses.
Note that GraVoS does not affect the inference time, since it is only applied at training. 

\figref{fig:gravos_gradient_mag} depicts the gradient magnitudes and the resulting $S^{mf}$ for a pre-trained detector (Voxel R-CNN~\cite{voxelrcnn}) for the three object classes of KITTI's benchmarks: {\em Car, Cyclist}, and {\em Pedestrian}. 
The background voxels have low gradient magnitudes (b), making them less likely than foreground voxels to be selected for training the detector (c). Conversely, many more points on the objects are maintained thanks to their high gradient magnitude. 
Almost all the voxels of the {\em Cyclist} and {\em Pedestrian} objects have high gradients, and  are therefore selected to the final subset, compared to the {\em Car} objects, with a somewhat smaller subset.

\begin{figure*}[t]
    \centering
    \includegraphics[width=0.9\linewidth]{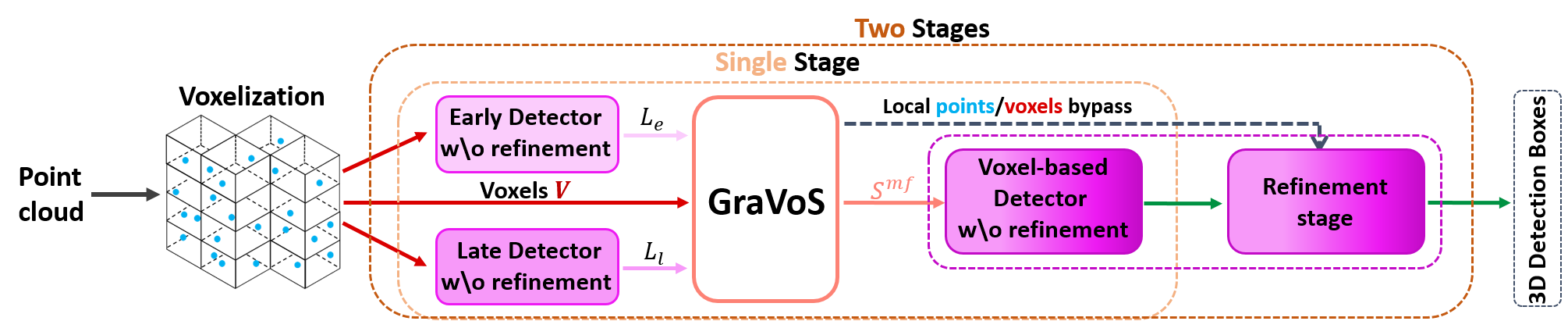}
    \caption{\textbf{Incorporating GraVoS into two-stage detectors}. 
    Voxel selection is performed during the first stage, as before.
    Since in two-stage architectures, the detector consists of a proposal generator and a refinement module, we use the detector without the refinement component in the first stage.
    The last refinement stage (in the second stage) gets the output of the proposal generator (the green arrow), as well as the required local data that is bypassed through GraVoS.
    }
    \label{fig:single_and_two_stages}
\end{figure*}

\input{tables/sota_comparison}
\input{tables/sota_comparison_bev}

\noindent
\textbf{GraVoS for two-stage detectors.} 
Generally speaking, 3D object detectors can be grouped by the number of stages in their detection pipeline, into single-stage~\cite{pointpillars,second,yang2018hdnet,yang2018pixor,yang20203dssd,yin2021center,zhou2018voxelnet} or two-stage detectors~\cite{mv3d,chen2019fast,voxelrcnn,ku2018joint,liang2019multi,qi2018frustum,pvrcnn,pointrcnn,parta2,wang2019frustum,Yang2018ipod,yang2019std}.
Single-stage approaches are fast since they usually have a single feed-forward network to predict the bounding boxes and classes. 
However, their main drawback is accuracy, since there is no component that specializes and fine tunes the box orientations. 
The two-stage approaches have an additional refinement module. 
This makes them slower and heavier in memory, but their accuracy is improved.

GraVoS is general in the sense that it can be used for both  single-stage and two-stage voxel-based detectors. 
However, the refinement stage (second stage) requires local information from the original input (point/voxels), which is not available after the selection process. 
As illustrated in \figref{fig:single_and_two_stages}, in this case, we apply GraVoS only on the first stage and transfer the missing local information (from the late stage detector) to the refinement stage. 
In our experiments (Section~\ref{sec:experiments}), we show the benefits of GraVoS for both single stage \cite{second,yin2021center} and two stage  \cite{voxelrcnn,parta2} approaches.

\section{Experiments}
\label{sec:experiments}
To evaluate the performance of our method, in Section~\ref{subsec:results} we compare  SoTA 3D object detection methods on the well established KITTI dataset~\cite{kitti2}, with and without our GraVoS module and training procedure. 
In addition, in Section~\ref{subsec:ablation} we explore several design choices for GraVoS.

\noindent
\textbf{KITTI dataset.}
KITTI~\cite{kitti2} is the most widely-used 3D object detection dataset for autonomous driving.
Its training set contains $7481$ examples that are divided into a training subset, with $3712$ examples, and a validation subset, with $3769$ examples~\cite{chen20153d}. 
The test set contains $7518$ examples.
We report results for all three classes in KITTI's benchmarks, \textit{Car}, \textit{Pedestrian} and \textit{Cyclist}, which contain $28742$, $4487$, and $1627$ object instances, respectively. 

\noindent
\textbf{Evaluation metrics.}
We report our results using the corrected average precision ($AP$) metric of \eqnref{eq:AP}, which is the standard for evaluating 3D detection~\cite{simonelli2019disentangling}:
\begin{equation}
    AP\lvert_R = \frac{1}{\abs{R}}\sum_{r \in R}\max_{r': r' \geq r} \rho(r').
    \label{eq:AP}
\end{equation}
Here, $\rho(r)$ is the precision at recall $r$, $R=[1/40,~ 2/40,~ \dots,~1]$ and $\abs{R}=40$. 
For precision and recall we use the standard IoU thresholds of $0.7, 0.5, 0.5$ for the \textit{Car}, \textit{Pedestrian}, and \textit{Cyclist} classes, respectively.

The evaluation on KITTI is divided into three difficulty categories: {\em Easy}, {\em Moderate} and {\em Hard}, based on the occlusion, the truncation and the object's distance from the scanner. 
The more distant and more occluded an object is, the harder it is to be detected.

\subsection{ Results}
\label{subsec:results}
To demonstrate the generality and the effectiveness of our method, we show that continuing to train four prominent 3D voxel-based object detectors with GraVoS selection yields improved performance on the challenging classes.
The four detectors are SECOND~\cite{second}, Voxel R-CNN~\cite{voxelrcnn}, Part-$A^2$~\cite{parta2} and CenterPoint~\cite{yin2021center}.
For each of these methods, we use its own protocol of augmentation.

\tabref{tab:sota_comparison} reports the results for the 3D detection benchmark and \tabref{tab:sota_comparison_bev} reports the results for the Bird-Eye View (BEV) detection benchmark.
GraVoS improves the overall performance of all the detectors.
Furthermore, GraVoS is especially beneficial for the non-prevalent classes i.e., {\em Cyclist} and {\em Pedestrian}, but might slightly degrade the results of the prevalent class {\em Car}.
This is attributed to the selection process, which prefers the non-prevalent classes.

\subsection{Ablation Study}
\label{subsec:ablation}
This section studies alternatives to GraVoS choices. 
In the following experiments, SECOND~\cite{second} is used as the baseline 3D object detector.

\begin{figure}[t]
    \centering
	\includegraphics[width=.76\linewidth]{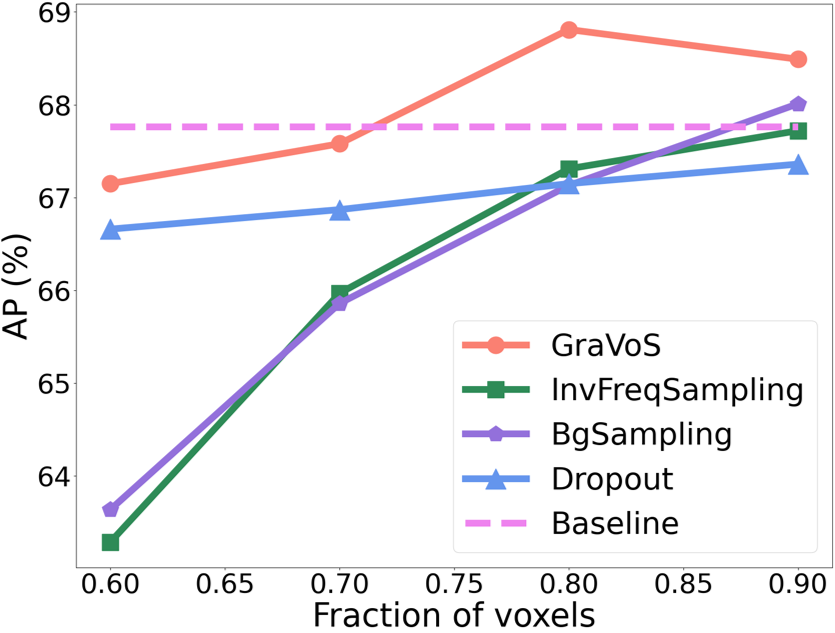}
	\caption{ 
    {\bf Comparison to alternative approaches.}
	GraVoS is compared to Dropout, BgSampling and InvFreqSampling  for different voxel selection ratios. 
	The baseline is the constant performance of the detector (all voxels).
	When too few voxels are used ($<0.7$), the detector misses objects, as expected.
	For ratios larger than $0.7$, GraVoS outperforms other approaches significantly.
	This is due to the fact that we use the meaningful voxels.
    }
	\label{fig:expLvs_ablation}
\end{figure}

\vspace{0.02in}
\noindent
\textbf{Voxel-selection alternatives.} 
\figref{fig:expLvs_ablation} shows that GraVoS significantly outperforms three alternatives:  Dropout, BgSampling and InvFreqSampling. 
{\em Dropout}  randomly selects a subset of the input voxels~\cite{srivastava2014dropout,cheng2020improving,choi2021part}. 
It was shown to avoid over-fitting and to improve the performance for many tasks. 
However, in our particular case, where the number of object voxels is significantly smaller than the total number of scene voxels, the benefit of Dropout is limited.
This is so since too few voxels that belong to object classes remain when sampling randomly.
See supplemental for additional results.

Interestingly, we further show that our performance is even better than sampling when the ground-truth voxel classes are known (which contradicts our assumption).
We utilized two classical class sampling techniques, in order to sample more voxels from the foreground (and from less prevalent classes) and less from the background.
In {\em BgSampling} voxels are randomly removed from the background until a threshold is met; 
in {\em InvFreqSampling}, sampling is based on the inverse of the class frequencies in a given scene.

\begin{figure}[t]
    \centering
    \begin{tabular}{cc}
    \subfloat{\includegraphics[width=0.44\linewidth]{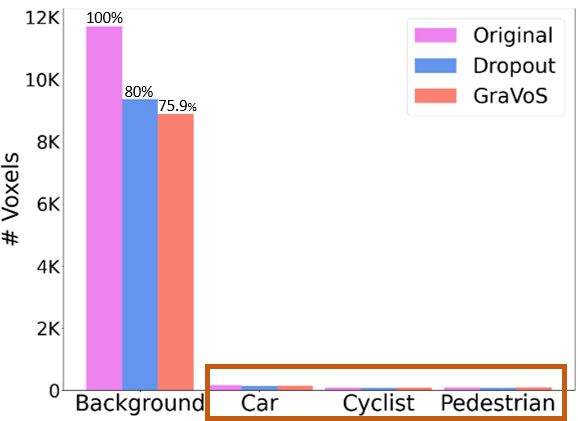}} &
    \subfloat{\includegraphics[width=0.44\linewidth]{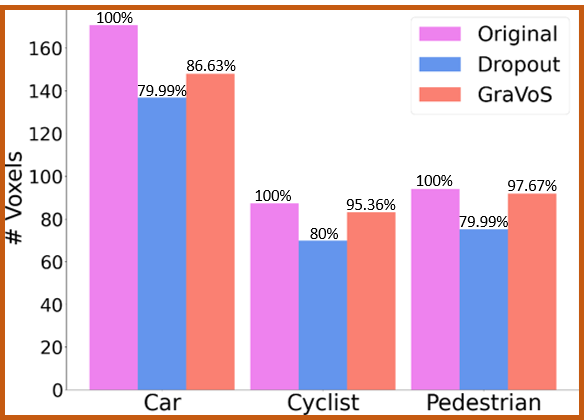}}\\
    \small (a) background \& foreground & \small (b) foreground zoom in
    \end{tabular}
    \caption{
        \textbf{GraVoS as a data balancer.}
        This figure shows the average number of voxels with and w/o GraVoS.
        (a)~The background has significantly more voxels than the foreground and accordingly, significantly higher voxel reduction.
        (b)~Zooming into the foreground, the {\em Car} category receives most of the voxel reduction.
    	Thus, GraVoS effectiveness may be attributed to data balancing.
    	}
	\label{fig:vox_balance_ablation}
\end{figure}

\vspace{0.02in}
\noindent
\textbf{Class-level voxel balancing with GraVoS.}
We analyze GraVoS's effectiveness by inspecting its influence on the average number of voxels for each object category. 
\figref{fig:vox_balance_ablation} shows that while GraVoS reduces the number of voxels in each object class, not all classes exhibit the same reduction. 
The background voxels are reduced by $24.1\%$, the {\em Car} category by $13.37\%$, while {\em Cyclist} and {\em Pedestrian} are hardly affected with only a $4.64\%$ and $2.33\%$ reduction, respectively. 
Essentially, GraVoS discards relatively more voxels from the background points than from objects, addressing the foreground-background imbalance challenge. 
This inherently differs from the naive Dropout, where the reduction is uniform across the whole scene ($20\%$).
This indicates that GraVoS has an object-level data balancing effect.

\vspace{0.02in}
\noindent
\textbf{Voxel selection ratio.} 
What shall be the number of selected voxels during training, $n_{vs}$? 
Since this parameter inherently depends on the number of input voxels, $n_v$, we explore the selection ratio
$\nu_{vs} = \frac{n_{vs}}{n_v}$. 
 \figref{fig:expLvs_ablation} shows that, as expected, it is beneficial to take a high ratio of voxels and that the best ratio is $0.8$.
This can be explained by the fact that uninformative voxels are discarded from the training process, allowing the network to focus on more informative voxels.
However, smaller ratios will cause a large gap between the distributions of the train and the test sets, which would result in an under performance of the detector.

\vspace{0.02in}
\noindent
\textbf{Intra-detector ratio.} 
We study the number of voxels that should be taken from each stage (early and late) of the pre-trained detector.
For that, we use the intra-detector ratio $\nu_{idr}$, which quantifies the fraction of voxels from the late stage detector w.r.t. the total number of selected voxels~$n_{vs}$.
Recall that when all the voxels are taken from the early-stage detector then $\nu_{idr} = 0$,  and when all the voxels are taken from the late-stage detector then $\nu_{idr} = 1$.
We start by setting $\nu_{idr} = 30/80$ and get class average accuracy of $67.58\%$.
Then, we increase the intra-detector ratio $\nu_{idr}$ by increments of $10/80$, taking more voxels from the late-stage than from the early-stage, up to $\nu_{idr} = 1$, which yields accuracy of $68.48\%$.
The best result is achieved for $\nu_{idr} = 50/80$, with detection accuracy of $68.81\%$.
(For this experiment we use a fixed voxel selection ratio $\nu_{vs}=0.8$.)

This study shows that selecting voxels mostly from the early-stage does not suffice for fine tuning the original detector (late-stage).
Moreover, it shows that the early-stage indeed provides additional information that the late-stage lacks, when a proper ratio is used.

\vspace{0.02in}
\noindent
\textbf{Early-stage detector's training duration.} 
In our framework, we consider the fully trained detector as the late-stage detector.
Therefore, we only have to choose for how long the detector needs to be trained in the early-stage.
To this end, we tested different epoch choices for the early-stage.
\tabref{tab:ablation_early_epochs} shows that the performance is in favor of earlier epochs, where the first epoch achieves the best result. 
This is expected since after the first epoch, the magnitude of the gradients at meaningful voxels with features that are easy to learn had not yet vanished. 
However, even for higher epochs using the early-stage is beneficial and achieves better results on average than the original detector ($67.76\%$).
(For this ablation study we used $\nu_{vs}=0.8$ and $\nu_{idr} = 50/80$.)

\begin{table}[]
    \centering
    \small
    \begin{tabular}{|c|c|c|c|c|c|c|}
    \hline
    Epoch & 1 & 5 & 10 & 20 & 30 & 40 \\
    \hline
    ${mAP_{3D}}$ & \textbf{68.81} & 68.59 & 68.59  & 68.58 & 68.34 & 68.27 \\
    \hline
    \end{tabular}
    \caption{\textbf{Training duration for the early-stage detector.} 
    ${mAP_{3D}}$ is the average precision over all the classes and difficulty levels for different early-stage detectors. 
    The best result is achieved after a single epoch.
    This may be attributed to the fact that the gradient magnitudes of easy-to-learn features are still non-negligible.}
    \label{tab:ablation_early_epochs}
    \vspace{-0.075in}
\end{table}

\vspace{0.02in}
\noindent
\textbf{Early and late detector mechanism choices.} 
We tested three different choices for the early and the late detectors: {\em mean}, {\em median}, and {\em top-k}.
For the mean and the median strategies, the voxels selected are those with gradients' magnitude higher than the mean or the median.
For the {\em top-k} strategy, we consider two choices for the intra-detector ratio, $\nu_{idr} = 50/80$ and $\nu_{idr} = 30/80$, where
$80\%$ of the total voxels were sampled in this experiment.

\tabref{tab:mechanisms_ablation} shows that most of the different choices improve the baseline detector~\cite{second}.
Selecting the {\em top-k} with $\nu_{idr} = 50/80$ for the late detector is the most beneficial, especially with the mean mechanism for the early detector.

\begin{table}[t]
    \centering
    \begin{tabular}{|c|c|c|}
     \hline
       Early mechanism & Late mechanism & ${mAP_{3D}}$ \\
    \hline
        \hline
        \multicolumn{2}{|c|}{Baseline~\cite{second}} & 67.76 \\
        \hline
        {\em mean} & {\em mean} & 68.25 \\
        {\em mean} & {\em median} & 68.01 \\
        {\em mean} & {\em top-k} ($\nu_{idr} = 50/80$) & \bf{68.81} \\
        {\em median} & {\em mean} & 67.80 \\
        {\em median} & {\em median} & 68.37 \\
        {\em median} & {\em top-k} ($\nu_{idr} = 50/80$) & 68.25 \\
        {\em top-k} ($\nu_{idr} = 50/80$) & {\em mean} & 67.33 \\
        {\em top-k} ($\nu_{idr} = 50/80$) & {\em median} & 68.07 \\
        {\em top-k} ($\nu_{idr} = 50/80$) & {\em top-k} ($\nu_{idr} = 30/80$) & 67.99\\
        {\em top-k} ($\nu_{idr} = 30/80$) & {\em top-k} ($\nu_{idr} = 50/80$) & 68.28 \\
    \hline
    \end{tabular}
    \caption{\textbf{Different mechanism choices for the detectors.} 
    Most selection choices for the early and late detectors improve the baseline detector.
    The best selection combination is the {\em top-k} with $\nu_{idr} = 50/80$ for the late detector and the mean selection for the early detector.
    In this experiment $80\%$ of the total voxels were sampled, inline with \figref{fig:expLvs_ablation}.
    }
    \label{tab:mechanisms_ablation}
    \vspace{-0.075in}
\end{table}

\vspace{0.02in}
\noindent
\textbf{Implementation details.}
We use the 3D detector implementations available in the OpenPCDet toolbox~\cite{openpcdet2020}. 
For a fair comparison, we use the default configurations for all detectors.
We set the voxel dimensions to be $(0.05, 0.05, 0.1)$, as provided in the toolbox.
For fine-tuning with GraVoS, we continue to train for $60$ epochs, $40$ epochs using the original detector's optimizer and $20$ epochs using stochastic gradient decent (SGD) and a step decay optimizer. 
Note that we use the same settings for all methods~\cite{second, voxelrcnn, parta2, yin2021center} after tuning for \cite{second} in our ablation study.
In the supplemental, we provide additional results that show that the performance boost is not attributed to longer training  or to further scheduler hyper-parameter details.
All experiments were done on a single NVIDIA $A100$ GPU.

\vspace{0.02in}
\noindent
\textbf{Limitations.} 
The main drawback of GraVoS is the need of further training, which means longer training times than those of the original detectors. 
Furthermore, during training, the memory and the computational requirements are higher, due to the  additional voxel selection stage. 
These limitations  apply only for the training stage. 
At inference, the memory and the time are the same as in the original detector.

\section{Conclusion}
This paper has presented a novel and generic voxel selection method---{\em Gradient-based Voxel Selection (GraVoS)}.
The key idea is to select voxels based on their meaningfulness to the detector.
GraVoS was shown to address two fundamental challenges in 3D detection datasets,  class-level data imbalance and foreground-background imbalance.

This paper has also proposed a training procedure that uses GraVoS to improve 3D detection without additional data.
This is done by utilizing the selected voxels exclusively for fine tuning the detector.

We have demonstrated that training four SoTA voxel-based detectors using our training approach and selected voxels yields a boost in performance.
The results are especially good when considering the challenging classes that have relatively few occurrences, regardless of difficulty.

An interesting future direction is to explore similar ideas of element selection on the raw cloud points. This will boost performance of detectors that do not use voxelization. 

\noindent
{\bf Acknowledgement.}
This work was supported by Advanced Defense Research Institute (ADRI)--Technion,  
and received funding from the European Union’s Horizon 2020 research and
innovation programme under the Marie Sklodowska-Curie grant agreement No 893465,
Israel Science Foundation 2329/22, and NVIDIA academic hardware grant.

\clearpage
{\small
\bibliographystyle{ieee_fullname}
\bibliography{references}
}

\end{document}

%% file: figures/gradients_fig.tex
\begin{figure*}[t]
\centering
\begin{tabular}{ccc}
\subfloat{\includegraphics[width=0.27\linewidth]{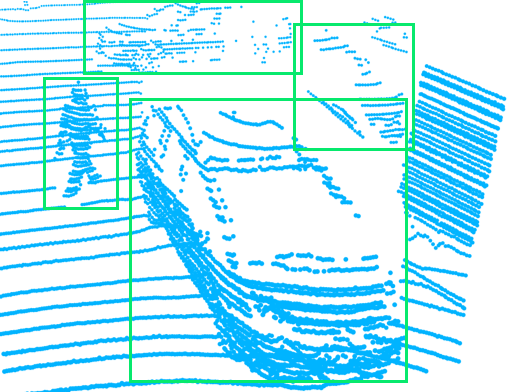}} &
\subfloat{\includegraphics[width=0.25\linewidth]{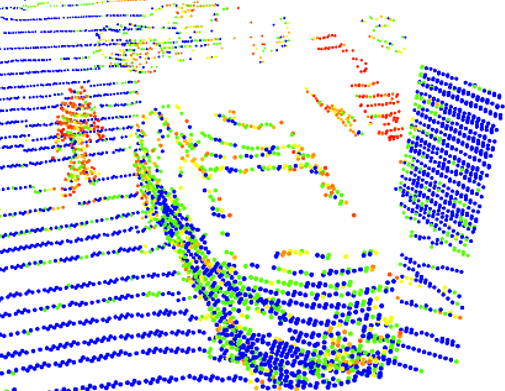}} &
\subfloat{\includegraphics[width=0.25\linewidth]{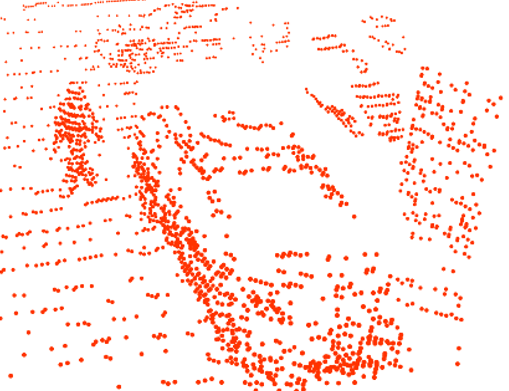}}\\
\subfloat{\includegraphics[width=0.27\linewidth]{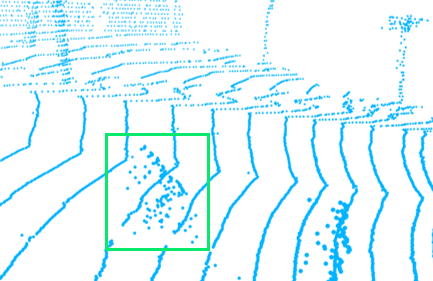}} &
\subfloat{\includegraphics[width=0.25\linewidth]{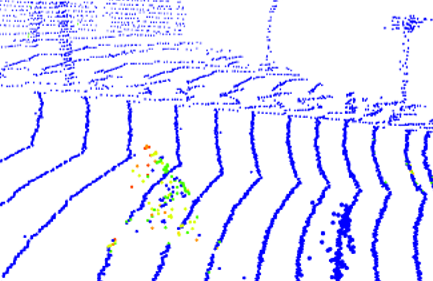}} &
\subfloat{\includegraphics[width=0.25\linewidth]{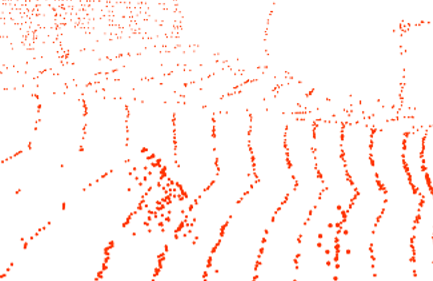}}\\
\subfloat{\includegraphics[width=0.27\linewidth]{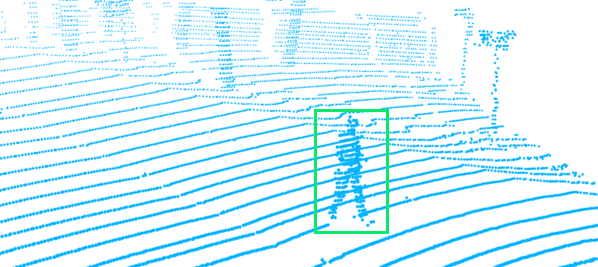}} &
\subfloat{\includegraphics[width=0.25\linewidth]{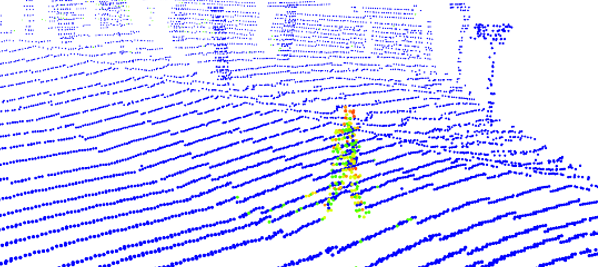}} &
\subfloat{\includegraphics[width=0.25\linewidth]{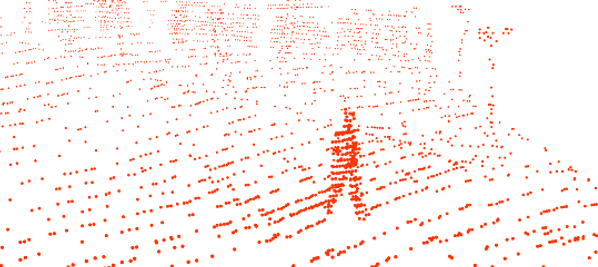}}\\
(a) Input (GT objects are in {\green rectangles}) & (b) Gradinent magnitudes & (c) Final subset $S^{mf}$ 
\end{tabular}
\caption{\textbf{Gradient-based voxel selection.}
Given an input {\cyan point cloud} (a), the magnitudes of the gradients are computed (b), and the {\salmon selected voxel subset} is computed (c).
The magnitude of the gradients is depicted as a colormap from blue to red (low to high values). 
Evidently, the gradients on the background voxels are lower and are therefore less likely to be selected than foreground pixels.
The objects' voxels have high gradients and therefore most of their voxels are retained in the final subset.
However, there are differences between the classes:
The less prominent classes, {\em Cyclist} (middle) and {\em Pedestrian} (bottom) retain relatively more points than the prominent class {\em Car} (top).
}
\label{fig:gravos_gradient_mag}
\end{figure*}

%% file: tables/sota_comparison.tex
\begin{table*}[t]
    \centering
    \small
    \begin{tabular}{|l |c c c| c c c| c c c| c c c c | c|}
    \hline
          & \multicolumn{3}{c}{\em Pedestrian}  & \multicolumn{3}{|c}{\em Cyclist} & \multicolumn{3}{|c|}{\em Car} & \multicolumn{4}{c|}{Average} \\
        Method & \multicolumn{1}{c}{Easy} & \multicolumn{1}{c}{Mod.} & \multicolumn{1}{c|}{Hard} & \multicolumn{1}{c}{Easy} & \multicolumn{1}{c}{Mod.} & \multicolumn{1}{c|}{Hard} & \multicolumn{1}{c}{Easy} & \multicolumn{1}{c}{Mod.} & \multicolumn{1}{c|}{Hard} &  
        \multicolumn{1}{c}{\em Ped.} & \multicolumn{1}{c}{\em Cyc.} & \multicolumn{1}{c}{\em Car}  & \multicolumn{1}{c|}{All} \\   
        \hline
    \hline
        Voxel R-CNN \cite{voxelrcnn} & 66.94 & 59.88 & 54.95 & 89.83 & 72.49 & \textbf{68.87} & \textbf{92.62} & 85.13 & 82.73 & 60.59 & 77.06 & 86.83 & 74.83 \\ 
        \cite{voxelrcnn}+Ours & \textbf{68.52} & \textbf{61.63} & \textbf{56.71} & \textbf{91.97} & \textbf{72.98} & 68.37  & 92.40 & \textbf{85.41} & \textbf{82.84} & \textbf{62.29} & \textbf{77.77} & \textbf{86.88} & \textbf{75.65} \\ 
        \hline
        SECOND \cite{second} & 54.90 & 49.84 & 45.15 & 81.85 & 65.42 & 61.26 & \textbf{90.79} & \textbf{81.87} & \textbf{78.75} & 49.96  & 69.51 & \textbf{83.80} & 67.76 \\
        \cite{second}+Ours & \textbf{57.75} & \textbf{51.99} & \textbf{47.54} & \textbf{84.36} & \textbf{66.41} & \textbf{62.61} & 89.53 & 81.06 & 78.07 & \textbf{52.43}  & \textbf{71.13} & 82.89 & \textbf{68.81} \\
        \hline
        Part-$A^2$ \cite{parta2} & 65.37 & 58.43 & 53.62 & 89.45 & 71.71 & 67.74 & \textbf{91.88} & \textbf{82.64} & 80.21 & 59.14 & 76.30 & 84.91 & 73.45 \\
        \cite{parta2}+Ours & \textbf{65.82} & \textbf{59.58} & \textbf{54.55} & \textbf{90.64} & \textbf{74.03} & \textbf{69.64} & 91.68 & 82.58 & \textbf{81.67} & \textbf{59.98} & \textbf{78.10} & \textbf{85.31} & \textbf{74.47} \\ 
        \hline
        CenterPoint \cite{yin2021center} & 56.85 & 53.17 & 49.73 & 80.27 & 62.85 & 60.13 & \textbf{89.58} & \textbf{82.09} & \textbf{79.58} & 53.25 & 67.75 & \textbf{83.75} & 68.25 \\
        \cite{yin2021center}+Ours & \textbf{58.02} & \textbf{54.64} & \textbf{50.94}  & \textbf{83.40} & \textbf{64.81} & \textbf{61.42} & 88.74 & 81.74 & 79.53 & \textbf{53.53} & \textbf{69.88} & 83.34 & \textbf{69.25} \\
        \hline
    \end{tabular}
\vspace{1pt}
\caption{
{\bf Performance on the 3D detection benchmark. }
Each method's performance is compared with and without our voxel selection. 
Results are reported for the Easy, Moderate (Mod.) and Hard categories on the three classes. 
Evidently, GraVoS always improves the average performance.
Furthermore, it improves the performance of the methods for the non-prevalent classes, while it might slightly degrade the performance for the prevalent class.
}
\label{tab:sota_comparison}
\end{table*}

%% file: tables/sota_comparison_bev.tex
\begin{table*}[t]
    \centering
    \small
    \begin{tabular}{|l |c c c| c c c| c c c| c c c c | c|}
    \hline
         & \multicolumn{3}{c}{\em Pedestrian}  & \multicolumn{3}{|c}{\em Cyclist} & \multicolumn{3}{|c|}{\em Car} & \multicolumn{4}{c|}{Average} \\
        Method & \multicolumn{1}{c}{Easy} & \multicolumn{1}{c}{Mod.} & \multicolumn{1}{c|}{Hard} & \multicolumn{1}{c}{Easy} & \multicolumn{1}{c}{Mod.} & \multicolumn{1}{c|}{Hard} & \multicolumn{1}{c}{Easy} & \multicolumn{1}{c}{Mod.} & \multicolumn{1}{c|}{Hard} &  
        \multicolumn{1}{c}{\em Ped.} & \multicolumn{1}{c}{\em Cyc.} & \multicolumn{1}{c}{\em Car}  & \multicolumn{1}{c|}{All} \\ \hline
        \hline
        Voxel R-CNN \cite{voxelrcnn} & 69.97 & 63.60 & 59.04 & 93.63 & 76.09 & \textbf{72.58} & \textbf{95.96} & 91.43 & \textbf{90.70} & 64.20 & 80.77 & \textbf{92.70} & 79.22 \\ 
        \cite{voxelrcnn}+Ours & \textbf{72.37} & \textbf{66.24} & \textbf{60.31} & \textbf{94.47} & \textbf{76.32} & 71.60 & \textbf{95.96} & \textbf{91.96} & 89.49 & \textbf{66.31} & \textbf{80.80} & 92.47 & \textbf{79.86} \\ 
        \hline
        SECOND \cite{second} & 60.94 & 55.73 & 51.56 & 87.87 & \textbf{70.91} & 66.57 & 92.30 & \textbf{89.68} & \textbf{87.51} & 56.08 & 75.12 & 89.83 & 73.67 \\
        \cite{second}+Ours & \textbf{62.23} & \textbf{56.78} & \textbf{52.63} & \textbf{89.02} & 70.88 & \textbf{66.77} & \textbf{92.86} & 89.62 & 87.26 & \textbf{57.21} & \textbf{75.56} & \textbf{89.91} & \textbf{74.23} \\ 
        \hline
        Part-$A^2$ \cite{parta2} & 68.31 & 61.70 & 57.33 & 91.19 & 75.42 & 70.97 & \textbf{92.89} & \textbf{90.14} & \textbf{88.17} & 62.45 & 79.19 & \textbf{90.40} & 77.35 \\
        \cite{parta2}+Ours & \textbf{68.51} & \textbf{62.40} & \textbf{58.04} & \textbf{93.13} & \textbf{75.91} & \textbf{72.68} & 92.85 & 90.07 & 88.13 & \textbf{62.98} & \textbf{80.57} & 90.35 & \textbf{77.97} \\ 
        \hline
        CenterPoint \cite{yin2021center} & 61.26 & 58.08 & 54.83 & 83.84 & 66.40 & 63.05 & \textbf{92.26} & \textbf{89.30} & \textbf{88.10} & 58.06 & 71.10 & \textbf{89.89} & 73.01 \\ 
        \cite{yin2021center}+Ours & \textbf{62.32} & \textbf{59.19} & \textbf{55.80} & \textbf{85.68} & \textbf{68.22} & \textbf{64.51} & 91.91 & 88.90 & 88.00 & \textbf{59.10} & \textbf{72.80} & 89.60 & \textbf{73.84} \\ 
        \hline
    \end{tabular}
\vspace{1pt}
\caption{{\bf Performance on the Bird-Eye View (BEV) detection  benchmark. }
As in \tabref{tab:sota_comparison}, our method is beneficial for all four methods.
}
\label{tab:sota_comparison_bev}
\end{table*}

%% file: GraVoS.bbl
\begin{thebibliography}{10}\itemsep=-1pt

\bibitem{chen20153d}
Xiaozhi Chen, Kaustav Kundu, Yukun Zhu, Andrew~G Berneshawi, Huimin Ma, Sanja Fidler, and Raquel Urtasun.
\newblock 3d object proposals for accurate object class detection.
\newblock {\em Advances in neural information processing systems}, 28, 2015.

\bibitem{mv3d}
Xiaozhi Chen, Huimin Ma, Ji Wan, Bo Li, and Tian Xia.
\newblock Multi-view 3d object detection network for autonomous driving.
\newblock In {\em Proceedings of the IEEE conference on Computer Vision and Pattern Recognition}, pages 1907--1915, 2017.

\bibitem{chen2019fast}
Yilun Chen, Shu Liu, Xiaoyong Shen, and Jiaya Jia.
\newblock Fast point r-cnn.
\newblock In {\em Proceedings of the IEEE/CVF International Conference on Computer Vision}, pages 9775--9784, 2019.

\bibitem{cheng2020improving}
Shuyang Cheng, Zhaoqi Leng, Ekin~Dogus Cubuk, Barret Zoph, Chunyan Bai, Jiquan Ngiam, Yang Song, Benjamin Caine, Vijay Vasudevan, Congcong Li, et~al.
\newblock Improving 3d object detection through progressive population based augmentation.
\newblock In {\em European Conference on Computer Vision}, pages 279--294. Springer, 2020.

\bibitem{choi2021part}
Jaeseok Choi, Yeji Song, and Nojun Kwak.
\newblock Part-aware data augmentation for 3d object detection in point cloud.
\newblock In {\em 2021 IEEE/RSJ International Conference on Intelligent Robots and Systems (IROS)}, pages 3391--3397. IEEE, 2021.

\bibitem{voxelrcnn}
Jiajun Deng, Shaoshuai Shi, Peiwei Li, Wengang Zhou, Yanyong Zhang, and Houqiang Li.
\newblock Voxel r-cnn: Towards high performance voxel-based 3d object detection.
\newblock {\em arXiv preprint arXiv:2012.15712}, 1(2):4, 2020.

\bibitem{kitti1}
Andreas Geiger, Philip Lenz, Christoph Stiller, and Raquel Urtasun.
\newblock Vision meets robotics: The kitti dataset.
\newblock {\em The International Journal of Robotics Research}, 32(11):1231--1237, 2013.

\bibitem{kitti2}
Andreas Geiger, Philip Lenz, and Raquel Urtasun.
\newblock Are we ready for autonomous driving? the kitti vision benchmark suite.
\newblock In {\em 2012 IEEE conference on computer vision and pattern recognition}, pages 3354--3361. IEEE, 2012.

\bibitem{sparseconv}
Benjamin Graham, Martin Engelcke, and Laurens Van Der~Maaten.
\newblock 3d semantic segmentation with submanifold sparse convolutional networks.
\newblock In {\em Proceedings of the IEEE conference on computer vision and pattern recognition}, pages 9224--9232, 2018.

\bibitem{graham2017submanifold}
Benjamin Graham and Laurens van~der Maaten.
\newblock Submanifold sparse convolutional networks.
\newblock {\em arXiv preprint arXiv:1706.01307}, 2017.

\bibitem{guo2020deep}
Yulan Guo, Hanyun Wang, Qingyong Hu, Hao Liu, Li Liu, and Mohammed Bennamoun.
\newblock Deep learning for 3d point clouds: A survey.
\newblock {\em IEEE transactions on pattern analysis and machine intelligence}, 43(12):4338--4364, 2020.

\bibitem{johnson2019survey}
Justin~M Johnson and Taghi~M Khoshgoftaar.
\newblock Survey on deep learning with class imbalance.
\newblock {\em Journal of Big Data}, 6(1):1--54, 2019.

\bibitem{ku2018joint}
Jason Ku, Melissa Mozifian, Jungwook Lee, Ali Harakeh, and Steven~L Waslander.
\newblock Joint 3d proposal generation and object detection from view aggregation.
\newblock In {\em 2018 IEEE/RSJ International Conference on Intelligent Robots and Systems (IROS)}, pages 1--8. IEEE, 2018.

\bibitem{pointpillars}
Alex~H Lang, Sourabh Vora, Holger Caesar, Lubing Zhou, Jiong Yang, and Oscar Beijbom.
\newblock Pointpillars: Fast encoders for object detection from point clouds.
\newblock In {\em Proceedings of the IEEE/CVF Conference on Computer Vision and Pattern Recognition}, pages 12697--12705, 2019.

\bibitem{liang2019multi}
Ming Liang, Bin Yang, Yun Chen, Rui Hu, and Raquel Urtasun.
\newblock Multi-task multi-sensor fusion for 3d object detection.
\newblock In {\em Proceedings of the IEEE/CVF Conference on Computer Vision and Pattern Recognition}, pages 7345--7353, 2019.

\bibitem{mitra2020multiview}
Rahul Mitra, Nitesh~B Gundavarapu, Abhishek Sharma, and Arjun Jain.
\newblock Multiview-consistent semi-supervised learning for 3d human pose estimation.
\newblock In {\em Proceedings of the ieee/cvf conference on computer vision and pattern recognition}, pages 6907--6916, 2020.

\bibitem{oksuz2020imbalance}
Kemal Oksuz, Baris~Can Cam, Sinan Kalkan, and Emre Akbas.
\newblock Imbalance problems in object detection: A review.
\newblock {\em IEEE transactions on pattern analysis and machine intelligence}, 43(10):3388--3415, 2020.

\bibitem{qi2018frustum}
Charles~R Qi, Wei Liu, Chenxia Wu, Hao Su, and Leonidas~J Guibas.
\newblock Frustum pointnets for 3d object detection from rgb-d data.
\newblock In {\em Proceedings of the IEEE conference on computer vision and pattern recognition}, pages 918--927, 2018.

\bibitem{qi2017pointnet}
Charles~R Qi, Hao Su, Kaichun Mo, and Leonidas~J Guibas.
\newblock Pointnet: Deep learning on point sets for 3d classification and segmentation.
\newblock In {\em Proceedings of the IEEE conference on computer vision and pattern recognition}, pages 652--660, 2017.

\bibitem{qi2017pointnet++}
Charles~Ruizhongtai Qi, Li Yi, Hao Su, and Leonidas~J Guibas.
\newblock Pointnet++: Deep hierarchical feature learning on point sets in a metric space.
\newblock {\em Advances in neural information processing systems}, 30, 2017.

\bibitem{ren2018learning}
Mengye Ren, Wenyuan Zeng, Bin Yang, and Raquel Urtasun.
\newblock Learning to reweight examples for robust deep learning.
\newblock In {\em International conference on machine learning}, pages 4334--4343. PMLR, 2018.

\bibitem{reuse2021ambiguity}
Matthias Reuse, Martin Simon, and Bernhard Sick.
\newblock About the ambiguity of data augmentation for 3d object detection in autonomous driving.
\newblock In {\em Proceedings of the IEEE/CVF International Conference on Computer Vision}, pages 979--987, 2021.

\bibitem{pvrcnn}
Shaoshuai Shi, Chaoxu Guo, Li Jiang, Zhe Wang, Jianping Shi, Xiaogang Wang, and Hongsheng Li.
\newblock Pv-rcnn: Point-voxel feature set abstraction for 3d object detection.
\newblock In {\em Proceedings of the IEEE/CVF Conference on Computer Vision and Pattern Recognition}, pages 10529--10538, 2020.

\bibitem{pointrcnn}
Shaoshuai Shi, Xiaogang Wang, and Hongsheng Li.
\newblock Pointrcnn: 3d object proposal generation and detection from point cloud.
\newblock In {\em Proceedings of the IEEE/CVF conference on computer vision and pattern recognition}, pages 770--779, 2019.

\bibitem{parta2}
Shaoshuai Shi, Zhe Wang, Jianping Shi, Xiaogang Wang, and Hongsheng Li.
\newblock From points to parts: 3d object detection from point cloud with part-aware and part-aggregation network.
\newblock {\em IEEE transactions on pattern analysis and machine intelligence}, 43(8):2647--2664, 2020.

\bibitem{shi2020point}
Weijing Shi and Raj Rajkumar.
\newblock Point-gnn: Graph neural network for 3d object detection in a point cloud.
\newblock In {\em Proceedings of the IEEE/CVF conference on computer vision and pattern recognition}, pages 1711--1719, 2020.

\bibitem{shrivastava2016training}
Abhinav Shrivastava, Abhinav Gupta, and Ross Girshick.
\newblock Training region-based object detectors with online hard example mining.
\newblock In {\em Proceedings of the IEEE conference on computer vision and pattern recognition}, pages 761--769, 2016.

\bibitem{simonelli2019disentangling}
Andrea Simonelli, Samuel~Rota Bulo, Lorenzo Porzi, Manuel L{\'o}pez-Antequera, and Peter Kontschieder.
\newblock Disentangling monocular 3d object detection.
\newblock In {\em Proceedings of the IEEE/CVF International Conference on Computer Vision}, pages 1991--1999, 2019.

\bibitem{srivastava2014dropout}
Nitish Srivastava, Geoffrey Hinton, Alex Krizhevsky, Ilya Sutskever, and Ruslan Salakhutdinov.
\newblock Dropout: a simple way to prevent neural networks from overfitting.
\newblock {\em The journal of machine learning research}, 15(1):1929--1958, 2014.

\bibitem{sung1996learning}
Kah-Kay Sung.
\newblock Learning and example selection for object and pattern detection.
\newblock 1996.

\bibitem{openpcdet2020}
OpenPCDet~Development Team.
\newblock Openpcdet: An open-source toolbox for 3d object detection from point clouds.
\newblock \url{https://github.com/open-mmlab/OpenPCDet}, 2020.

\bibitem{thrun2006stanley}
Sebastian Thrun, Mike Montemerlo, Hendrik Dahlkamp, David Stavens, Andrei Aron, James Diebel, Philip Fong, John Gale, Morgan Halpenny, Gabriel Hoffmann, et~al.
\newblock Stanley: The robot that won the darpa grand challenge.
\newblock {\em Journal of field Robotics}, 23(9):661--692, 2006.

\bibitem{wang2019frustum}
Zhixin Wang and Kui Jia.
\newblock Frustum convnet: Sliding frustums to aggregate local point-wise features for amodal 3d object detection.
\newblock In {\em 2019 IEEE/RSJ International Conference on Intelligent Robots and Systems (IROS)}, pages 1742--1749. IEEE, 2019.

\bibitem{second}
Yan Yan, Yuxing Mao, and Bo Li.
\newblock Second: Sparsely embedded convolutional detection.
\newblock {\em Sensors}, 18(10):3337, 2018.

\bibitem{yang2018hdnet}
Bin Yang, Ming Liang, and Raquel Urtasun.
\newblock Hdnet: Exploiting hd maps for 3d object detection.
\newblock In {\em Conference on Robot Learning}, pages 146--155. PMLR, 2018.

\bibitem{yang2018pixor}
Bin Yang, Wenjie Luo, and Raquel Urtasun.
\newblock Pixor: Real-time 3d object detection from point clouds.
\newblock In {\em Proceedings of the IEEE conference on Computer Vision and Pattern Recognition}, pages 7652--7660, 2018.

\bibitem{yang20203dssd}
Zetong Yang, Yanan Sun, Shu Liu, and Jiaya Jia.
\newblock 3dssd: Point-based 3d single stage object detector.
\newblock In {\em Proceedings of the IEEE/CVF conference on computer vision and pattern recognition}, pages 11040--11048, 2020.

\bibitem{Yang2018ipod}
Zetong Yang, Yanan Sun, Shu Liu, Xiaoyong Shen, and Jiaya Jia.
\newblock {IPOD:} intensive point-based object detector for point cloud.
\newblock {\em CoRR}, abs/1812.05276, 2018.

\bibitem{yang2019std}
Zetong Yang, Yanan Sun, Shu Liu, Xiaoyong Shen, and Jiaya Jia.
\newblock Std: Sparse-to-dense 3d object detector for point cloud.
\newblock In {\em Proceedings of the IEEE/CVF International Conference on Computer Vision}, pages 1951--1960, 2019.

\bibitem{ye2020hvnet}
Maosheng Ye, Shuangjie Xu, and Tongyi Cao.
\newblock Hvnet: Hybrid voxel network for lidar based 3d object detection.
\newblock In {\em Proceedings of the IEEE/CVF conference on computer vision and pattern recognition}, pages 1631--1640, 2020.

\bibitem{yin2021center}
Tianwei Yin, Xingyi Zhou, and Philipp Krahenbuhl.
\newblock Center-based 3d object detection and tracking.
\newblock In {\em Proceedings of the IEEE/CVF conference on computer vision and pattern recognition}, pages 11784--11793, 2021.

\bibitem{zheng2021cia}
Wu Zheng, Weiliang Tang, Sijin Chen, Li Jiang, and Chi-Wing Fu.
\newblock Cia-ssd: Confident iou-aware single-stage object detector from point cloud.
\newblock In {\em Proceedings of the AAAI conference on artificial intelligence}, volume~35, pages 3555--3562, 2021.

\bibitem{zheng2021se}
Wu Zheng, Weiliang Tang, Li Jiang, and Chi-Wing Fu.
\newblock Se-ssd: Self-ensembling single-stage object detector from point cloud.
\newblock In {\em Proceedings of the IEEE/CVF Conference on Computer Vision and Pattern Recognition}, pages 14494--14503, 2021.

\bibitem{zhou2018voxelnet}
Yin Zhou and Oncel Tuzel.
\newblock Voxelnet: End-to-end learning for point cloud based 3d object detection.
\newblock In {\em Proceedings of the IEEE conference on computer vision and pattern recognition}, pages 4490--4499, 2018.

\end{thebibliography}
